\pdfoutput=1

\documentclass[11pt]{article}

\usepackage[final]{coling}

\usepackage{times}
\usepackage{latexsym}

\usepackage[T1]{fontenc}

\usepackage[utf8]{inputenc}

\usepackage{microtype}

\usepackage{inconsolata}

\usepackage{graphicx}

\usepackage{lipsum}
\usepackage{booktabs}
\usepackage{xcolor}
\usepackage{enumitem}
\usepackage{comment}
\usepackage{caption}
\usepackage{subcaption}
\usepackage{listings}
\usepackage{color,colortbl}
\usepackage{todonotes} 
\usepackage{soul}
\usepackage{threeparttable, makecell}
\usepackage{makecell}
\usepackage{xcolor}
\usepackage{array}
\usepackage{tcolorbox}
\usepackage{multirow}

\title{GenAI Content Detection Task 3: Cross-Domain Machine-Generated Text Detection Challenge}

\author{Liam Dugan$^1$, \hspace{0.15cm} Andrew Zhu$^1$, \hspace{0.15cm} Firoj Alam$^2$, \hspace{0.15cm} Preslav Nakov$^3$\\
\textbf{Marianna Apidianaki}$^1$, \hspace{0.15cm} \textbf{Chris Callison-Burch}$^1$ \\
University of Pennsylvania$^1$ \hspace{0.25cm} Qatar Computing Research Institute$^2$ \hspace{0.25cm} MBZUAI$^3$\\
{\tt \normalsize \{ldugan, andrz, marapi, ccb\}@seas.upenn.edu}\\ \tt \normalsize preslav.nakov@mbzuai.ac.ae, fialam@hbku.edu.qa}

\newcommand{\leidos}{[{\color{orange}Le}]}
\newcommand{\pangram}{[{\color{red}Pa}]}
\newcommand{\mosaic}{[{\color{violet}Mo}]}
\newcommand{\cnlp}{[{\color{magenta}Cn}]}
\newcommand{\oneeight}{[{\color{purple}80}]}
\newcommand{\random}{[{\color{cyan}Ra}]}
\newcommand{\luxveri}{[{\color{teal}Lx}]}
\newcommand{\alert}{[{\color{olive}Al}]}
\newcommand{\ustc}{[{\color{brown}Us}]}
\newcommand{\baseline}{[{\color{blue}Ba}]}

\begin{document}
\maketitle
\begin{abstract}
Recently there have been many shared tasks targeting the detection of generated text from Large Language Models (LLMs). However, these shared tasks tend to focus either on cases where text is limited to one particular domain or cases where text can be from many domains, some of which may not be seen during test time. In this shared task, using the newly released RAID benchmark, we aim to answer whether or not models can detect generated text from a large, yet fixed, number of domains and LLMs, all of which are seen during training. Over the course of three months, our task was attempted by 9 teams with 23 detector submissions. We find that multiple participants were able to obtain accuracies of over 99\% on machine-generated text from RAID while maintaining a 5\% False Positive Rate---suggesting that detectors are able to robustly detect text from many domains and models simultaneously. We discuss potential interpretations of this result and provide directions for future research.
\end{abstract}

\section{Introduction}
The detection of AI generated text is an increasingly relevant task in the modern age. Such detection can help combat misinformation \cite{sharevski2023talking}, phishing attacks \cite{bethany2024largelanguagemodellateral}, and other fraudulent activities \cite{weiss2019,lund2023chatgpt}. This is particularly important given that humans struggle to detect generated text reliably \cite{dugan-etal-2020-roft,dugan-etal-2023-roft} and that generated text is often more persuasive than human-written text \cite{spitale-etal-2023-disinformation}.

Recently there has been an increase in large scale shared tasks for AI generated text detectors \cite{wang-etal-2024-semeval-2024,Fivez_etal_2024,bevendorff-etal-2024}. These shared tasks tend to either focus on one particular domain or hold out many domains and LLM generators in the final test set. This prevents us from understanding how well a single detector can detect text from a large fixed set of models and domains. Such a setting is important to understand as it helps us to delineate the boundaries of our detector capabilities. For example, it is clear that detectors trained on a single LLM can accurately detect text from that model \cite{solaiman-etal-2019} but does this extend to arbitrarily many LLMs? What about arbitrarily many domains?

In particular we attempt to answer the following research questions:
\begin{itemize}
\item \textit{(RQ1)} \textit{Can a single detection model be trained to detect generated text from many different known domains and LLMs accurately?}
\item \textit{(RQ2)} \textit{Can a single detection model be robust to many different known adversarial attacks?}
\end{itemize}

In order to test these research questions, we conduct this shared task using the newly released RAID benchmark \cite{dugan-etal-2024-raid}. RAID is a dataset of over 10 million documents from 11 generative models, 8 textual domains, 4 decoding strategies, and 11 adversarial attacks. We chose RAID as it is one of the largest benchmarks currently available, and it features variation across decoding strategies and adversarial attacks as well as a large variety of textual domains. Crucially, the test set of RAID does not include any held-out models or domains and has yet to be released publicly. Therefore RAID allows us to answer our research questions most effectively.

We received 23 submissions to the shared task from 9 different teams, and 7 system description papers. 
The results of the evaluation are publicly available.\footnote{\texttt{https://raid-bench.xyz/shared-task}} The two best performing teams (Pangram and Leidos) achieved extremely strong performance without adversarial attacks (99.3\%) and with adversarial attacks (97.7\%). In this paper we will summarize our general takeaways from this result and offer future directions for effective benchmarking of generated text classifiers.

\section{Related Work}
The shared tasks most similar to ours are the SemEval24 Task 8 and GenAI Content Detection Task 1 \cite{wang-etal-2024-semeval-2024,wang2025genai}. Both of these tasks include outputs from many generative models and domains in both their train and test set. However, as mentioned in the introduction, these tasks hold out many domains and models in their test set to test the generalization performance of classifiers to new unseen models and domains. In our task we explicitly give all domain and model information to our participants up front and only hold out particular articles within such domains.

\begin{table}[t]
\centering 
\small
\begin{tabular}{p{0.20\linewidth}|p{0.68\linewidth}} 
\toprule
\multicolumn{2}{c}{\textbf{Sources in the RAID dataset}}\\
\midrule
\textbf{Generative Models} ($n=11$)&GPT2, GPT3, GPT4, Cohere, Cohere Chat, MPT 30B, MPT 30B Chat, Mistral 7B, Mistral 7B Chat, ChatGPT, Llama 70B Chat\\
\midrule
\textbf{Domains} ($n=8$)&Abstracts, Recipes, Books, Reddit, News, Reviews, Poetry, Wiki\\
\midrule
\textbf{Decoding Strategies} ($n=4$)&Greedy (temp=0), Random Sampling (temp=1, top-p=1), Greedy + Repetition Penalty, Sampling + Repetition Penalty\\
\midrule
\textbf{Adversarial Attacks} ($n=11$)&Alternative Spelling, Homoglyph, Article Deletion, Number Swap, Insert Paragraphs, Upper Lower Swap, Paraphrase, Synonym Swap, Zero Width Space, Misspelling, Whitespace Addition\\
\bottomrule
\end{tabular}
\caption{The domains, models, decoding strategies and attacks covered by RAID.}
\label{tab:raid-categories}
\end{table}

Other shared tasks in the past have also evaluated generated text detectors in specific high-risk domains such as academic essays \cite{llm-detect-ai-generated-text,chowdhury2025genai} scientific papers \cite{kashnitsky-etal-2022-overview} or news articles \cite{bevendorff-etal-2024}. Previous shared tasks have also studied detection in a multilingual context \cite{Shamardina_2022,sarvazyan2023overviewautextificationiberlef2023,Fivez_etal_2024,wang2025genai,chowdhury2025genai}. While such tasks are interesting and valuable, they do not test what we are interested in, namely the ability of single detectors to detect text from many different models and domains.

Finally, the recent Voight-Kampff task \cite{bevendorff-etal-2024} is particularly noteworthy. In their task they employ a set of builders (who build detectors) and breakers (who create adversarial datasets to fool the detectors). They are the first shared task to explicitly include adversarial constraints into their evaluation---experimenting with homoglyph attacks and different detailed prompt formulations with bullet points and summaries of the original source texts. However, they conduct their task in a pairwise manner, giving each detector two texts, one of which must be human-written, and asking the detector to select the human-written text. While they showed that detectors can do well on this highly adversarial task (96.1 ROC AUC score for the top team), we target the more difficult yet more realistic version of the task, where a single document is given as input.

\begin{table}[t]
    \small
    \centering
    \begin{tabular}{l|c|c|c|c|c}
    \toprule
    &\textbf{Num.}&&\textbf{Self-}&\textbf{PPL}&\textbf{PPL}\\
    \textbf{Model}&\textbf{Gens}&\textbf{Toks}&\textbf{BLEU}&\textbf{-L7B}&\textbf{-G2X}\\
    \midrule
    \textbf{Human}&14971&378.5&7.64&9.09&21.2\\
    \midrule
    \textbf{GPT 2}&59884&384.7&23.9&8.33&8.10\\
    \textbf{GPT 3}&29942&185.6&13.6&3.90&8.12\\
    \textbf{ChatGPT}&29942&329.4&10.3&3.39&9.31\\
    \textbf{GPT 4}&29942&350.8&9.42&5.01&13.4\\
    \textbf{Cohere}&29942&301.9&11.0&5.67&23.7\\
    \textit{(+ Chat)}&29942&239.0&11.0&4.93&11.6\\
    \textbf{Mistral}&59884&370.2&19.1&7.74&17.9\\
    \textit{(+ Chat)}&59884&287.7&9.16&4.31&10.3\\
    \textbf{MPT}&59884&379.2&22.1&14.0&66.9\\
    \textit{(+ Chat)}&59884&219.2&5.39&7.06&56.3\\
    \textbf{LLaMA}&59884&404.4&10.6&3.33&9.76\\
    \midrule
    \textbf{Total}&509k&323.4&13.7&6.61&23.8\\
    \bottomrule
    \end{tabular}
    \caption{Statistics for the generations in train and test without adversarial attacks. \textbf{PPL-L7B} refers to mean perplexity according to LLaMA 7B and \textbf{PPL-G2X} refers to mean perplexity according to GPT 2 XL.}
    \label{tab:dataset-statistics}
\end{table}

\begin{table*}[t]
    \setlength{\tabcolsep}{3pt}
    \small
    \centering
    \begin{tabular}{cl|c|ccccccccccc}
    \toprule
    &&Human&ChatGPT&dav.-003&GPT-4&Cohere&Coh.-C&GPT-2&MPT&MPT-C&Mistral&Mist.-C&Llama2-C\\
    \midrule
    \parbox[t]{2mm}{\multirow{8}{*}{\rotatebox[origin=c]{90}{\textbf{Train}}}}&Abstracts & 1766 & 3532 & 3532 & 3532 & 3532 & 3532 & 7064 & 7064 & 7064 & 7064 & 7064 & 7064\\
    &Books & 1781 & 3562 & 3562 & 3562 & 3562 & 3562 & 7124 & 7124 & 7124 & 7124 & 7124 & 7124\\
    &News & 1780 & 3560 & 3560 & 3560 & 3560 & 3560 & 7120 & 7120 & 7120 & 7120 & 7120 & 7120\\
    &Poetry & 1771 & 3542 & 3542 & 3542 & 3542 & 3542 & 7084 & 7084 & 7084 & 7084 & 7084 & 7084\\
    &Recipes & 1772 & 3544 & 3544 & 3544 & 3544 & 3544 & 7088 & 7088 & 7088 & 7088 & 7088 & 7088\\
    &Reddit & 1779 & 3558 & 3558 & 3558 & 3558 & 3558 & 7116 & 7116 & 7116 & 7116 & 7116 & 7116\\
    &Wiki & 1779 & 3558 & 3558 & 3558 & 3558 & 3558 & 7116 & 7116 & 7116 & 7116 & 7116 & 7116\\
    &Reviews & 943 & 1886 & 1886 & 1886 & 1886 & 1886 & 3772 & 3772 & 3772 & 3772 & 3772 & 3772\\
    \midrule
    &Total & 13371 & 26742 & 26742 & 26742 & 26742 & 26742 & 53484 & 53484 & 53484 & 53484 & 53484 & 53484\\
    \midrule
    \parbox[t]{2mm}{\multirow{8}{*}{\rotatebox[origin=c]{90}{\textbf{Test}}}}&Abstracts & 200 & 400 & 400 & 400 & 400 & 400 & 800 & 800 & 800 & 800 & 800 & 800\\
    &Books& 200 & 400 & 400 & 400 & 400 & 400 & 800 & 800 & 800 & 800 & 800 & 800\\
    &News& 200 & 400 & 400 & 400 & 400 & 400 & 800 & 800 & 800 & 800 & 800 & 800\\
    &Poetry& 200 & 400 & 400 & 400 & 400 & 400 & 800 & 800 & 800 & 800 & 800 & 800\\
    &Recipes& 200 & 400 & 400 & 400 & 400 & 400 & 800 & 800 & 800 & 800 & 800 & 800\\
    &Reddit& 200 & 400 & 400 & 400 & 400 & 400 & 800 & 800 & 800 & 800 & 800 & 800\\
    &Wiki& 200 & 400 & 400 & 400 & 400 & 400 & 800 & 800 & 800 & 800 & 800 & 800\\
    &Reviews& 200 & 400 & 400 & 400 & 400 & 400 & 800 & 800 & 800 & 800 & 800 & 800\\
    \midrule
    &Total & 1600 & 6400 & 3200 & 3200 & 3200 & 3200 & 3200 & 6400 & 6400 & 6400 & 6400 & 6400\\
    \bottomrule
    \end{tabular}
    \caption{Number of documents in the RAID dataset by model and domain. Each human-written document has exactly one machine-written counterpart from each model with each of the decoding strategies listed in Table \ref{tab:raid-categories}. Due to lack of support for repetition penalty sampling, API-based models have two outputs per human document and open-weight models have four outputs per human document. ``-C'' in model name indicates the chat fine-tuned version of the model. The adversarial data has an identical distribution but with 12x more documents per cell.}
    \label{tab:raid_source_counts}
\end{table*}

\section{Task Setup}
\subsection{RAID Benchmark}
The RAID benchmark was created by sampling roughly 2000 human-written documents from each of 8 domains. Then, for each human-written document, a machine-written version is generated from each of the 11 LLMs with each of the 4 decoding strategies. Finally, each of the 11 adversarial attacks are applied to all machine-written documents. The test set consists of 200 human-written documents per domain selected from the same distribution as the train set and all generations based on those documents. The documents were then checked to prevent duplication and leakage and to ensure no overlap between train and test data.

In Table \ref{tab:raid-categories}, we list the domains, models, decoding strategies, and adversarial attacks covered in RAID. In Table \ref{tab:dataset-statistics} we report summary statistics and in Table \ref{tab:raid_source_counts} we report the exact distribution of documents from the training set and test set.

\subsection{Subtask A: Cross-Domain Detection}
For Subtask A, participants were asked to submit detectors that would be robust to all 8 domains in the main RAID dataset without adversarial attacks.
In Table \ref{tab:domains}, we provide a breakdown of the addressed domains, and links to the original data sources for extra training. We provided our teams with these links in order to help them secure as much training data as possible. This is particularly important given that RAID has a roughly 40:1 class imbalance of AI vs. human-written text. We observed that many teams took advantage of this and sampled extra human data from these sources.

\subsection{Subtask B: Adversarial Robustness}
In Subtask B, the participants had  to evaluate their detectors on all data from Subtask A with the addition of 11 adversarial attacks. In Table \ref{tab:adversarial} we list the adversarial attacks applied as well as the source papers that first study them. For this subtask, the participants also had access to the original code\footnote{\texttt{https://github.com/liamdugan/raid/tree/main/ generation/adversarial/attackers}} used to create the adversarial attacks. This allowed them to adversarially modify any existing piece of data to assist in training.

\subsection{Baselines}
We use the following models as baselines:
\begin{itemize}[noitemsep]
    \item \textbf{Openai-RoBERTa-large} \cite{solaiman-etal-2019}:  RoBERTa-large model fine-tuned on GPT-2 generations.
    \item \textbf{RADAR} \cite{hu-etal-2023-radar}: Vicuna 7B model trained on adversarial paraphrases
    \item \textbf{Binoculars} \cite{hans2024spotting}: Current SoTA metric-based detection model. Uses perplexity divided by cross-entropy between Falcon 7B and Falcon 7B Instruct. 
    \item \textbf{GLTR} \cite{gehrmann-etal-2019-gltr}: Baseline metric-based detection model. Uses GPT-2 small and rank=10 for vocabulary cutoff.
\end{itemize}
We gave our participants access to the code to run these baselines as well as the outputs for these models on the training set.

\section{Metrics}
\subsection{Performance Metric}
Due to the class imbalance in the RAID dataset, typical metrics like Accuracy and AUROC are not appropriate for this task. Thus, in keeping with the original RAID paper, we use domain-adjusted TPR@FPR=5\% as our metric. This metric represents how much of the generated text we are able to correctly identify while maintaining a 5\% false positive rate (where a false positive is defined as incorrectly labeling a human-written text as being machine generated).

\subsection{Threshold Search}
In order to measure TPR@FPR=5\% we need to find a binary classification threshold that results in a 5\% false positive rate on the human data for each detector and domain. The search algorithm we use for this purpose is the same algorithm that is described in the RAID paper. We start at the threshold corresponding to the mean score of human data (50\% FPR), and approach the desired false positive rate by iteratively incrementing or decrementing the threshold. If we overshoot the target FPR, we divide our increment in half and flip the sign. We continue to do this process until the false positive rate is within $\epsilon=0.0005$ of the desired false positive rate or until 50 iterations are reached. If 50 iterations are reached without convergence, then we select the threshold corresponding to the FPR that is closest to the target while still being less than the target.

\subsection{Robustness Metric}
In addition to this performance metric, we also calculate the standard deviation of TPR@FPR=5\%. This measures how robust each model is across each domain of comparison. For subtask A, this metric will be measured across domains and for subtask B, this metric will be measured across adversarial attacks.

\section{Submissions}
We received 23 submissions to the shared task from 9 different teams, and system description papers from 7 of 9 teams. In this section, we describe the systems submitted by each team in detail. In Section \ref{sec:results} we will discuss aggregate results for the teams and in Section \ref{sec:trends} we will discuss the broader trends across our participant submissions.

\paragraph{Team LuxVeri \luxveri} \cite{genai-detect:2025:task:LuxVeri}: This team fine-tuned both RoBERTa-base and RoBERTa-base-openai-detector on a subset of the RAID training data for 3 epochs, using a learning rate of 2e-5, a batch size of 4, and the AdamW optimizer. These trained models were then used to compute weights for ensembling based on an inverse perplexity weighting technique, which was applied to the ensemble for the adversarial task. For the non-adversarial task, they only used RoBERTa-base with the same hyperparameters.

\paragraph{Team Random \random} \cite{genai-detect:2025:task:Random}: This team's contributions include a pipeline that integrates XLM-RoBERTa embeddings for enhanced text representation, domain adaptation using a Domain-Adversarial Neural Network (DANN) \cite{ganin-etal-dann} to minimize domain-specific biases and improve generalization across diverse text domains. They also incorporate adversarial attack classification to detect and mitigate manipulative techniques.

\paragraph{Team USTC-BUPT \ustc} This team fine-tuned RoBERTa-Large via focal loss \cite{lin-etal-focal-loss} on the RAID training set by adding four samples for each human sample using synonym replacement. They then down-sampled the AI-written texts from a 10:1 ratio to a 2:1 ratio of AI to human text to form the training set. In the hyperparameters for focal loss, they set alpha to 0.65 and gamma to 2.5. In addition to this, they also had a secondary submission where they simply fine-tuned RoBERTa-base on a subset of the RAID training set.

\begin{table*}[t]
    \small
    \centering
    \begin{tabular}{l|c|c|c|c|c|c|c|c|c}
    \toprule 
    \multicolumn{10}{c}{\textbf{Subtask A: Performance Across Domains (Official Results)}}\\
    \midrule
    &\textbf{News}&\textbf{Wiki}&\textbf{Reddit}&\textbf{Books}&\textbf{Abs.}&\textbf{Reviews}&\textbf{Poetry}&\textbf{Recipes}&\textbf{Total} ($\sigma$)\\
    \midrule
    \leidos ~Leidos v1.0.3&\textbf{99.9}&99.8&98.3&99.4&99.9&98.6&99.3&\textbf{100.0}&\textbf{99.4} (0.6)\\
    \pangram ~Pangram&99.7&99.1&98.5&99.5&99.3&\textbf{99.6}&98.8&99.9&99.3 (\textbf{0.4})\\
    \leidos ~Leidos v1.0.2&\textbf{99.9}&\textbf{99.9}&\textbf{99.4}&99.5&99.9&95.9&\textbf{99.6}&\textbf{100.0}&99.3 (1.2)\\
    \leidos ~Leidos v1.0.4&\textbf{99.9}&99.7&99.0&99.3&\textbf{100.0}&96.5&99.4&\textbf{100.0}&99.2 (1.1)\\
    \leidos ~Leidos v1.0.1&\textbf{99.9}&99.8&98.6&99.4&99.9&96.2&99.4&\textbf{100.0}&99.1 (1.2)\\
    \ustc ~R-L Focal Loss&99.0&97.8&96.1&98.1&99.8&97.0&97.0&99.9&98.1 (1.3)\\
    \alert ~ALERT v1.1&99.7&95.4&75.7&\textbf{99.9}&99.9&87.2&78.3&98.3&91.8 (9.4)\\
    \cnlp ~DistilBERT-NITS&89.9&87.7&90.0&93.5&90.9&85.9&90.0&96.0&90.5 (2.9)\\
    \alert ~ALERT v1.2&99.5&91.3&87.2&99.2&99.9&89.9&64.9&82.8&89.3 (11.0)\\
    \luxveri ~R-B \& R-Oai&87.5&90.2&62.4&89.5&99.2&83.7&73.5&75.1&82.6 (10.9)\\
    \luxveri ~R-Oai \& BERT&91.8&87.3&75.1&87.1&97.0&86.0&76.3&59.4&82.5 (11.1)\\
    \luxveri ~Fine-tuned R-B&87.5&89.7&61.7&89.6&98.8&82.5&66.3&74.6&81.3 (11.9)\\
    \baseline ~Binoculars&80.7&76.5&81.3&82.8&76.0&78.0&80.1&76.4&79.0 (2.4)\\
    \mosaic ~MOSAIC-4&79.5&67.6&78.2&79.8&77.1&81.4&63.7&75.8&75.2 (5.9)\\
    \mosaic ~MOSAIC-5&79.0&65.8&76.7&79.8&76.5&77.2&64.8&75.1&74.5 (5.4)\\
    \luxveri ~Radar \& R-L&91.6&73.7&76.3&78.1&74.2&58.7&45.7&73.5&71.5 (12.8)\\
    \baseline ~RADAR&87.4&77.3&73.6&78.1&67.5&6.3&46.0&88.7&65.6 (25.7)\\
    \baseline ~GLTR&67.7&63.6&63.2&71.9&60.1&65.0&18.2&67.9&59.7 (16.0)\\
    \oneeight ~L3-60 Zero-shot&63.6&36.5&61.5&65.4&55.3&68.9&51.5&53.9&57.1 (9.6)\\
    \oneeight ~M3-60 Zero-shot&58.1&58.1&65.8&63.3&44.1&67.1&53.2&50.5&56.5 (7.4)\\
    \baseline ~openai-roberta-large&67.8&59.4&60.0&52.5&64.8&52.8&23.3&65.1&55.7 (13.3)\\
    \cnlp ~Adv.-submission-3&27.1&26.1&52.8&57.1&30.1&48.6&38.0&94.0&46.7 (21.1)\\
    \cnlp ~Adv.-New-Detector&14.0&16.2&40.4&39.2&34.7&29.4&17.8&91.0&35.3 (23.2)\\
    \ustc ~Roberta\_dataaug.&4.6&3.6&40.5&7.3&83.1&3.1&5.1&98.8&30.8 (36.8)\\
    \cnlp ~Adv.\_Data\_Detector&10.1&17.5&27.9&24.8&27.7&28.7&13.5&88.0&29.8 (23.0)\\
    \luxveri ~Radar R-B CGPT-R&20.0&16.0&4.8&2.5&51.1&62.1&4.4&32.9&24.2 (21.1)\\
    \random ~Adv. CDMGTD&4.2&3.4&2.1&2.1&6.8&2.9&1.7&2.4&3.2 (1.6)\\
    \midrule
    Average Performance&70.7&66.6&68.4&71.8&74.6&67.7&58.1&78.5&69.5 (5.7)\\
    \bottomrule
    \end{tabular}
    \caption{TPR at FPR=5\% for detectors across different domains on the RAID test set along with their standard deviation ($\sigma$). Baselines are given the \baseline  ~tag. ``Abs.'' is shorthand for Abstracts. Team rankings are determined based on the highest performing submission for each team (see Table \ref{tab:team_ranking}).}
    \label{tab:per_domain_results}
\end{table*}

\paragraph{Team ALERT \alert} \cite{genai-detect:2025:task:bbn-uo-alert}: This team's approach uses authorship style representations to distinguish between human-authored and machine-generated text across various domains. Their authorship attribution (AA) systems are trained with contrastive learning. They employ an ensemble-based AA system (ALERT v1.1) that integrates stylistic embeddings from two complementary subsystems: One system that focuses on cross-genre robustness with hard positive and negative mining strategies using Linq-Embed-Mistral \cite{LinqAIResearch2024} as the backbone architecture, and a second system with Semantic, Lexical, Clustering based hard positive and negative mining which uses Qwen2-1.5B.

\paragraph{Team Leidos \leidos} \cite{genai-detect:2025:task:Leidos}: This team trained four Distil-RoBERTa-Base detectors to evaluate both binary and multi-class classification. They also explored the effects of class weighting to address dataset imbalance. The ``v1.0.1'' detector is a binary classifier without class weighting, the ``v1.0.3'' detector is a binary classifier with class weighting, the ``v1.0.4'' is a multi-class classifier without class weighting and the ``v1.0.2'' detector is a multi-class classifier with class weighting.

\paragraph{Team MOSAIC \mosaic} \cite{genai-detect:2025:task:MOSAIC}: This team submitted a completely unsupervised approach which uses a mixture of models to score the texts. Their ensemble method is grounded in fundamental information-theoretic principles from universal compression in order to optimally combine the strengths of multiple LLMs for machine-generated text detection. The four models used are Tower-7b, Tower-13b, Llama-2-7b and Llama-2-7b-chat.

\begin{table*}[t]
    \small
    \centering
    \begin{tabular}{l|c|c|c|c|c|c|c|c|c|c|c|c}
    \toprule 
    \multicolumn{13}{c}{\textbf{Subtask B: Performance Across Adversarial Attacks (Official Results)}}\\
    \midrule
    &\textbf{AS}&\textbf{AD}&\textbf{HG}&\textbf{IP}&\textbf{NS}&\textbf{PP}&\textbf{MS}&\textbf{SY}&\textbf{UL}&\textbf{WS}&\textbf{ZW}&\textbf{Total} ($\sigma$)\\
    \midrule
    \leidos ~Leidos v1.0.2&99.2&99.0&\textbf{97.3}&98.7&99.2&92.3&98.8&98.6&98.9&99.0&92.7&\textbf{97.7} (\textbf{2.5})\\
    \pangram ~Pangram&99.2&98.7&91.9&\textbf{99.3}&99.2&91.6&99.0&96.2&\textbf{99.3}&\textbf{99.3}&99.3&\textbf{97.7} (2.9)\\
    \leidos ~Leidos v1.0.4&99.1&99.0&94.7&98.7&99.2&94.8&98.8&98.6&98.9&98.8&90.9&97.6 (2.6)\\
    \leidos ~Leidos v1.0.3&\textbf{99.3}&\textbf{99.3}&93.6&98.7&\textbf{99.4}&\textbf{96.3}&\textbf{99.2}&\textbf{99.1}&99.2&99.2&84.2&97.2 (4.4)\\
    \leidos ~Leidos v1.0.1&99.0&99.0&86.1&98.1&99.1&94.8&98.9&98.8&98.8&98.5&78.8&95.7 (6.4)\\
    \ustc ~R-L Focal Loss&97.9&98.2&84.5&93.6&98.1&84.0&97.8&97.4&97.9&97.9&67.1&92.7 (9.5)\\
    \alert ~ALERT v1.1&91.8&92.1&68.5&89.7&91.8&57.7&91.0&87.3&91.3&91.2&46.8&82.6 (15.5)\\
    \luxveri ~Fine-tuned R-B&80.5&78.1&90.4&79.8&79.8&77.9&77.9&74.4&75.0&66.2&\textbf{100.0}&80.1 (8.4)\\
    \alert ~ALERT v1.2&89.9&89.0&61.9&84.1&88.6&57.1&88.6&84.1&87.2&85.6&40.2&78.8 (16.0)\\
    \luxveri ~R-B \& R-Oai&81.7&79.4&41.7&81.2&81.1&78.1&79.3&75.8&76.1&68.0&86.9&76.0 (11.6)\\
    \luxveri ~R-Oai \& BERT&81.6&79.4&20.9&81.7&82.2&75.8&79.6&77.6&76.7&77.1&83.7&74.9 (17.0)\\
    \baseline ~Binoculars&78.2&74.3&37.7&71.7&77.1&80.3&78.0&43.5&73.8&70.1&99.1&71.3 (16.2)\\
    \baseline ~Radar&70.8&67.9&59.3&73.7&71.0&67.3&69.5&67.5&70.4&66.1&82.2&69.6 (5.3)\\
    \mosaic ~MOSAIC-5&72.2&69.5&90.2&73.3&69.7&70.3&71.7&22.7&66.5&67.0&85.5&69.4 (16.3)\\
    \mosaic ~MOSAIC-4&72.9&70.8&86.6&74.5&71.3&71.9&72.5&28.5&68.6&67.5&71.4&69.3 (13.6)\\
    \luxveri ~Radar \& R-L&70.3&61.2&21.2&73.0&69.9&73.0&63.9&74.9&55.7&60.2&91.3&65.5 (16.6)\\
    \baseline ~GLTR&61.2&52.1&24.3&61.4&59.9&47.2&59.8&31.2&48.1&45.8&97.2&53.5 (18.1)\\
    \oneeight ~L3-60 Zero-shot&56.6&50.5&3.0&57.4&56.3&50.6&55.6&53.5&57.1&61.9&57.1&51.4 (15.4)\\
    \baseline ~openai-roberta-L&52.4&33.2&21.3&55.1&51.7&72.9&39.5&79.4&19.3&40.1&99.9&51.3 (23.6)\\
    \oneeight ~M3-60 Zero-shot&55.6&48.6&3.6&56.7&52.2&37.7&53.7&40.2&56.5&59.7&56.5&48.1 (15.4)\\
    \cnlp ~Adv.-sub.-3&46.7&45.1&20.8&46.7&46.5&18.0&46.8&41.6&46.7&46.7&46.7&41.6 (10.4)\\
    \cnlp ~Adv.-New-Det.&35.3&35.2&18.9&35.3&35.4&11.9&35.4&31.6&35.3&35.3&35.3&31.7 (7.7)\\
    \ustc ~Roberta\_dataaug.&30.8&31.6&16.4&31.8&30.8&26.8&30.4&30.1&30.8&29.5&11.6&27.6 (6.5)\\
    \cnlp ~Adv.\_Data\_Det.&29.7&29.4&18.5&29.8&29.6&8.5&29.8&26.9&29.8&29.8&29.8&26.8 (6.5)\\
    \luxveri ~Radar R-B C-R&22.3&15.2&0.4&4.9&22.0&34.9&18.1&30.0&6.6&4.3&11.0&16.2 (10.6)\\
    \random ~Adv. CDMGTD&3.2&3.0&24.8&3.2&3.2&3.5&3.2&3.2&3.2&3.2&20.8&6.5 (7.6)\\
    \midrule
    Average Performance&68.4&65.3&49.2&67.4&67.9&60.6&66.8&61.3&64.1&64.2&67.9&64.3 (5.3)\\
    \bottomrule
    \end{tabular}
    \caption{TPR at FPR=5\% for detectors across different adversarial attacks along with their standard deviation ($\sigma$). Baselines are given the \baseline ~tag. Abbreviations are: AS: Alternative Spelling, AD: Article Deletion, HG: Homoglyph, IP: Insert Paragraphs, NS: Number Swap, PP: Paraphrase, MS: Misspelling, SY: Synonym Swap, UL: Upper Lower Swap, WS: Whitespace Addition, ZW: Zero-Width Space Addition. Team rankings determined by the highest performing submission (see Table \ref{tab:team_ranking_adversarial}).}
    \label{tab:adversarial_results}
\end{table*}

\paragraph{Team CNLP-NITS-PP \cnlp} \cite{genai-detect:2025:task:CNLP-NITS-PP}: This team submitted a model that first classifies whether the text has been adversarially attacked or not. If an attack is detected, the text undergoes preprocessing to mitigate the attack, after which the preprocessed text proceeds to the  model for MGT detection. Finally, their detector works by fine-tuning a DistilBERT model to extract semantic features. 

\paragraph{Team 1-800-SHARED-TASKS \oneeight} \cite{kadiyala-2024-rkadiyala,ram_kadiyala_2024}: This team submitted a model that uses a token classifier for detecting change in authorship within the same text. Their models were not trained on RAID and so all data from the test set is out-of-distribution. In addition, inference was done without any pre-processing against adversarial methods, nor were those methods present in the training data. The final generated score was based on the proportion of tokens classified as AI generated among the ones within the supported context length.

\paragraph{Team Pangram \pangram} \cite{genai-detect:2025:Task3:Pangram}: This team pretrained an autoregressive LLM-based detector on a wide variety of datasets, domains, languages, prompt schemes, and LLMs used to generate the AI portion of the dataset. They aggressively employed several augmentation strategies and preprocessing strategies to improve robustness. They then mined the RAID train set for the AI examples with the largest error based on the original classifier,  mixed those examples and their human-written counterparts back into the training set, and retrained the detector until convergence.

\section{Results}
\label{sec:results}

\subsection{Subtask A: Cross-Domain Detection}
In Table \ref{tab:per_domain_results}, we report the official results for Subtask A. Overall, we see that a large fraction of our teams beat our provided baselines and established strong new results. The winning team for subtask A, Leidos, achieved 99.4\% TPR across all 8 domains---a substantial improvement over the prior state of the art result on RAID. This suggests that it is possible to build classifiers that are robust across a finite set of domains and models.

While the top teams get consistently high accuracy, this does not mean that all domains are equally difficult. In terms of comparisons between domains, the most difficult domain to classify when averaged across all submissions was Poetry (58.1\%), and the easiest were Recipes (78.5\%) and Abstracts (74.6\%). AI-generated Poems may be more difficult to identify as they rely on rare and unusual word choice, and could exhibit low overall likelihood even to well trained classifiers. On the other hand, both the Recipes and Abstracts domains were unexpectedly easy for the detectors, potentially due to the highly formulaic nature of these domains.

\subsection{Subtask B: Adversarial Robustness}
In Table \ref{tab:adversarial_results}, we report the official results for subtask B. We observe similarly high performance, with the two winning teams for subtask B (Leidos and Pangram) getting a surprisingly high 97.7\% TPR on the dataset. Once again, we see many teams beat our strong baselines and feel that these results validate our intuition that classifiers, when given a finite set of adversarial attacks to defend against, can do quite well.

In addition, while many attacks from the original RAID paper are effective when the defender is not prepared for them (e.g.,  Whitespace Insertion, Article Deletion), such attacks seem to be relatively easy to defend against. One interesting finding is that the most difficult attacks to defend against were the Homoglyph attack (49.2\%), the Paraphrase attack (60.6\%), and the Synonym attack (61.3\%). Despite having access to the code that generated the Homoglyph attack, it is still difficult to create a model that is robust to this attack without any text preprocessing or normalization. 

In addition, while many attacks are easy to defend against with good knowledge of the attacker, it seems that both the Paraphrase and Synonym attacks remain difficult to deal with even when given access to the attack generation code. This is likely because these attacks are not easily solved with simple preprocessing techniques, and because they require models to learn alternative distributions that are often fairly distinct from generic generated text.

\section{Broader Trends}
\label{sec:trends}
Across all submissions to the shared task we find the following broader trends:

\paragraph{Trend 1: Text Preprocessing} The first clear trend we saw was the use of text preprocessing. Team \cnlp ~and~\random ~both trained classifiers to detect particular adversarial attacks in the dataset in order to apply preprocessing, and team \pangram  ~simply ran all incoming text to the detector through text normalization. Looking at the results, these methods seemed to be largely effective at neutralizing many of the simpler adversarial attacks such as Zero-Width Space, Upper-Lower Swap, Insert Paragraphs, Whitespace Insertion and Alternative Spelling. However, other attacks which either create or destroy vital information in the text such as Paraphrasing, Synonym Swap, Article Deletion, and Misspelling seem to be more difficult to preprocess away. Such attacks may require other forms of prevention that go beyond text normalization.

\paragraph{Trend 2: Hard Positive and Negative Sampling} The second clear trend we found was that many teams looked explicitly for harder examples in the dataset. Of the top four teams, three reported using this particular technique. Team \ustc ~used focal loss, which concentrates learning on hard misclassified examples; Team \alert ~used pairs of difficult examples to train their contrastive learning objective, and Team \pangram ~looked specifically for examples where their classifier had large error and incorporated them into their existing pre-training dataset. While it is difficult to isolate the effect of any one method or technique in a study like this, it is clear that sampling particularly hard examples is a promising direction for building robust detectors.

\paragraph{Trend 3: Diversity of Approach} The third and final trend we observe is the diversity of approaches we witnessed. We saw submissions involving unsupervised methods \mosaic, ensemble methods \luxveri, token-level models \oneeight, contrastive learning models \alert, and multi-class classification methods \leidos. Each of the submissions differed not only in terms of performance, but also in adversarial and cross-domain robustness properties. For example, we see that the unsupervised approaches \mosaic~exhibited lower overall performance but higher cross-domain robustness than ensemble methods \luxveri. This suggests that there is a large potential for novel modeling work to be done in detection. 

\section{Conclusion}
In real-world scenarios, API-based detectors can expect a majority of their text to come from a relatively small set of LLMs and domains. The most common frontier models (Gemini Pro, Claude 3.5, GPT-4o) will mostly likely be used to generate text in a set of high-risk domains (academic essays, news articles, scientific papers)  with potentially some paraphrasing or synonym changes applied to help avoid detection. In such cases, it is desirable for a detector to focus on optimizing performance as much as possible on this fixed set of models and domains while not caring as much about performance on other less common models. Our results show that detectors can exhibit strong performance in such constrained settings.

\begin{table*}
\centering 
\small
\begin{tabular}{p{0.11\linewidth}|p{0.83\linewidth}} 
\toprule
\multicolumn{2}{p{0.94\linewidth}}{\textbf{Domain}: Abstracts, \textbf{Title}: EdgeFlow: Achieving Practical Interactive Segmentation with Edge-Guided Flow}\\
\midrule
\textbf{ChatGPT}&EdgeFlow is a novel approach to interactive image segmentation that combines edge detection and flow-based methods to achieve practical and efficient results. The proposed method utilizes an edge-guided flow algorithm to guide the segmentation process, allowing users to interactively refine the segmentation boundaries. The algorithm incorporates both local and global information to accurately capture [...]\\
\midrule
\textbf{Human}&High-quality training data play a key role in image segmentation tasks.

Usually, pixel-level annotations are expensive, laborious and time-consuming

for the large volume of training data. To reduce labelling cost and improve

segmentation quality, interactive segmentation methods have been proposed, [...]\\
\midrule
\multicolumn{2}{p{0.94\linewidth}}{\textbf{Domain}: Recipes, \textbf{Title}: Olive Spirals}\\
\midrule
\textbf{ChatGPT}&Instructions:

1. Preheat your oven to 400°F (200°C) and line a baking sheet with parchment paper.

2. In a bowl, combine the black olives, green olives, sun-dried tomatoes, feta cheese, basil, parsley, olive oil, salt, and pepper. Mix well to combine all the ingredients.

3. Roll out the puff pastry sheet on a lightly floured surface into a rectangle shape, about 1/4 inch thick.

4. Spread the olive mixture evenly over the puff pastry, leaving a small border around the edges. [...]\\
\midrule
\textbf{Human}&Mix yeast, sugar\& 125ml warm water in a bowl. Cover and set aside in a warm place for 10 minutes, or until frothy. Sift the flour and salt into a bowl and make a well in the centre. Add frothy yeast, oil and 250ml of warm water. Mix to a soft dough and gather into a ball. Turn out on a floured surface and knead for 10 minutes until smooth. Cover loosely with greased plastic wrap and set aside for 1 hour until [...]\\
\midrule
\end{tabular}
\caption{Comparison of human-written text to chatgpt-written text in the Abstracts and Recipes domain from the RAID dataset. We can see that the human-written abstract has periodic newline characters and the generated text does not. In addition we see that the ChatGPT-written recipe has numbered lists of instructions while the human-written text is written in paragraph form. Artifacts such as these may trivialize the detection task.}
\label{tab:confounds}
\end{table*}

Future work should seek to further target this particular setting and replicate these findings on a much more diverse corpus. Such a study would entail collecting many generations from a core set of models in high risk domains and conduct a heavy adversarial attack circuit. Since prompt-based attacks are likely to be common (e.g. “write in a way that is not detectable”) these should be a major focus. In addition, testing a large set of “humanizing” paraphrase models is also desirable. It is unclear whether or not our results will extend to models that have significant prompt and paraphrase variations applied.

Finally, it is worth noting that detectors doing well in constrained settings does not imply generalization to unseen models. Detectors still suffer from poor generalization across unseen models and domains as discussed in \citet{dugan-etal-2024-raid}. However, this discrepancy between in-distribution and out-of-distribution performance is something that is worth highlighting as a potential source for future investigation.

\section*{Limitations}

\paragraph{Test Data Leakage} The first thing to note is the potential for test data leakage. Since the human-written documents are sampled from publicly available datasets and we give our participants the links to such datasets, there is a chance that participants models have seen the human-written documents in our test data and have overfit to them, allowing for high performance. In our eyes this is unlikely, as most of the linked datasets have a large amount of documents and the percentage of documents included in test data from those is vanishingly small. In addition, such leakage would only be problematic when searching for thresholds. Since our metric is TPR, we only measure the accuracy of the classifier in identifying machine-written text, all of which is hidden and has never been released publicly. Nonetheless, this is an important caveat to include.

\paragraph{Confounding Factors} As documented by \citet{gritsai2025aidetectorsgoodenough} many datasets for AI-detection exhibit consistent confounding features that trivialize the detection task. On manual investigation of the RAID dataset we found instances of such confounding factors in the data that would potentially make detection easier (Table \ref{tab:confounds}). For example, all human-written recipes were written without numbered lists of steps whereas all generated recipes included numbered lists of steps. To investigate the effects of these confounds we manually cleaned the data to remove the most egregious examples and trained a RoBERTa-base model on both the original and cleaned RAID. We saw a drop from 92.67 to 89.67 after cleaning the data---a small but significant performance difference. We are continuing to investigate this and do not yet have enough evidence to conclude whether or not this is the source of the high performance. We hope that future studies can help to shed more light on this issue and establish good quality standards for benchmark datasets.

\section*{Acknowledgments}
This research is supported in part by the Office of the Director of National Intelligence (ODNI), Intelligence Advanced Research Projects Activity (IARPA), via the HIATUS Program contract \#2022-22072200005. The views and conclusions contained herein are those of the authors and should not be interpreted as necessarily representing the official policies, either expressed or implied, of ODNI, IARPA, or the U.S. Government. The U.S. Government is authorized to reproduce and distribute reprints for governmental purposes notwithstanding any copyright annotation therein.

\bibliography{custom}

\appendix

\section{Team Rankings}
Teams are ranked by their highest performing submission. In Table \ref{tab:team_ranking} we report the official team ranking for Subtask A and in Table \ref{tab:team_ranking_adversarial} we report the official ranking for Subtask B

\section{Extra RAID Details}
In Table \ref{tab:domains} we list the 8 domains present in RAID along with clickable links to the human-written sources from which the data was sampled. These links were given to the participants to assist in curating extra data for training. 

In Table \ref{tab:adversarial} we list the 11 adversarial attacks applied to the RAID data along with the relative percentage of attack surface used for the attack and the papers each attack originally came from. We provided participants with the code for these attacks to allow them to train on arbitrarily many examples of each attack at varying attack strengths.

\section{Recommendations for Future Evaluations}
In this section, we will outline recommendations for future robustness studies and shared tasks. These recommendations come not only from our experiences with conducting this shared task, but also from discussions we had with participants about potential areas for future improvements.

\begin{table}[t]
    \small
    \centering
    \begin{tabular}{l|c|c}
    \toprule 
    \multicolumn{3}{c}{\textbf{Team Ranking (Subtask A)}}\\
    \midrule
    &\textbf{Best Submission}&\textbf{Result}\\
    \midrule
    \leidos ~Leidos&Leidos v1.0.3&\textbf{99.4} (0.6)\\
    \pangram ~Pangram&Pangram&99.3 (\textbf{0.4})\\
    \ustc ~USTC&R-L Focal Loss&98.1 (1.3)\\
    \alert ~ALERT&ALERT v1.1&91.8 (9.4)\\
    \cnlp ~CNLP&DistilBERT-NITS&90.5 (2.9)\\
    \luxveri ~LuxVeri&R-B \& R-Oai&82.6 (10.9)\\
    \midrule
    \baseline ~\textbf{Baseline}&Binoculars&79.0 (2.4)\\
    \midrule
    \mosaic ~MOSAIC&MOSAIC-4&75.2 (5.9)\\
    \oneeight ~1-800&L3-60 Zero-shot&57.1 (9.6)\\
    \random ~Random&Adv. CDMGTD&3.2 (1.6)\\
    \bottomrule
    \end{tabular}
    \caption{Team ranking for Subtask A ranked by their best submission. Metric used for result is TPR at FPR=5\% along with their standard deviation ($\sigma$). See Table \ref{tab:per_domain_results} for full results.}
    \label{tab:team_ranking}
\end{table}

\begin{table}[t]
    \small
    \centering
    \begin{tabular}{l|c|c}
    \toprule 
    \multicolumn{3}{c}{\textbf{Team Ranking (Subtask B)}}\\
    \midrule
    &\textbf{Best Submission}&\textbf{Result}\\
    \midrule
    \leidos ~Leidos&Leidos v1.0.2&\textbf{97.7} (\textbf{2.5})\\
    \pangram ~Pangram&Pangram&\textbf{97.7} (2.9)\\
    \ustc ~USTC&R-L Focal Loss&92.7 (9.5)\\
    \alert ~ALERT&ALERT v1.1&82.6 (15.5)\\
    \luxveri ~LuxVeri&Fine-tuned R-B&80.1 (8.4)\\
    \midrule
    \baseline ~\textbf{Baseline}&Binoculars&71.3 (16.2)\\
    \midrule
    \mosaic ~MOSAIC&MOSAIC-5&69.4 (16.3)\\
    \oneeight ~1-800&L3-60 Zero-shot&51.4 (15.4)\\
    \cnlp ~CNLP&Adv.-sub.-3&41.6 (10.4)\\
    \random ~Random&Adv. CDMGTD&6.5 (7.6)\\
    \bottomrule
    \end{tabular}
    \caption{Team ranking for Subtask B ranked by their best submission. Metric used for result is TPR at FPR=5\% along with their standard deviation ($\sigma$). See Table \ref{tab:adversarial_results} for full results.}
    \label{tab:team_ranking_adversarial}
\end{table}

\begin{table}
\centering 
\small
\begin{tabular}{p{0.17\linewidth}|p{0.28\linewidth}|p{0.34\linewidth}} 
\toprule
\textbf{Domain}&\textbf{Source}&\textbf{Description}\\
\midrule
\textbf{Abstracts}&\href{https://www.kaggle.com/datasets/Cornell-University/arxiv}{arxiv.org}&ArXiv Abstracts\\
\textbf{Recipes}&\href{https://recipenlg.cs.put.poznan.pl/}{allrecipes.com}&Ingredients + Recipe\\
\textbf{Books}&\href{https://paperswithcode.com/dataset/cmu-book-summary-dataset}{wikipedia.org}&Plot Summaries\\
\textbf{Reddit}&\href{https://huggingface.co/datasets/sentence-transformers/reddit-title-body}{reddit.com}&Reddit Posts\\
\textbf{News}&\href{https://github.com/derekgreene/bbc-datasets}{bbc.com/news}&News Articles\\
\textbf{Reviews}&\href{https://ieee-dataport.org/open-access/imdb-movie-reviews-dataset}{imbd.com}&Movie Reviews\\
\textbf{Poetry}&\href{https://www.kaggle.com/datasets/michaelarman/poemsdataset}{poemhunter.com}&Poems (Any Style)\\
\textbf{Wiki}&\href{https://huggingface.co/datasets/aadityaubhat/GPT-wiki-intro}{wikipedia.org}&Article Introductions\\
\bottomrule
\end{tabular}
\caption{All domains in the RAID dataset alongside a description of where they are from. Clickable source links go directly to the source dataset from which the human samples were taken.}
\label{tab:domains}
\end{table}

\begin{table}[t]
\centering 
\small
\begin{tabular}{p{0.35\linewidth}|r|p{0.4\linewidth}} 
\toprule
\textbf{Attack}&\textbf{$\theta$}&\textbf{Source}\\
\midrule
\textbf{Alternative Spelling}&100\%&\cite{liang2023gpt}\\
\textbf{Article Deletion}&50\%&\cite{liang2023mutationbased,guerrero2022mutationbased}\\
\textbf{Homoglyph}&100\%&\cite{wolff2020attacking,gagiano-etal-2021-robustness}\\
\textbf{Insert Paragraphs}&50\%&\cite{bhat-parthasarathy-2020-effectively}\\
\textbf{Number Swap}&50\%&\cite{bhat-parthasarathy-2020-effectively}\\
\textbf{Paraphrase}&100\%&\cite{krishna2023paraphrasing,sadasivan2023aigenerated}\\
\textbf{Misspelling}&20\%&\cite{liang2023mutationbased,gagiano-etal-2021-robustness,gao2018blackbox}\\
\textbf{Synonym}&50\%&\cite{pu-etal-2023-deepfake}\\
\textbf{Upper Lower}&5\%&\cite{gagiano-etal-2021-robustness}\\
\textbf{Whitespace}&20\%&\cite{cai2023evade,gagiano-etal-2021-robustness}\\
\textbf{Zero-Width Space}&100\%&\cite{guerrero2022mutationbased}\\
\bottomrule
\end{tabular}
\caption{The adversarial attacks used in the project. $\theta$ represents the manually determined fraction of available attacks carried out. We determine this fraction through manual review.}
\label{tab:adversarial}
\end{table}

\paragraph{Create a common preprocessing script.} This should be done with the explicit goal of removing any and all potential confounding factors that do not have to do with the text itself. We suggest making the preprocessing script public so that participants can apply it to existing pre-training data and any other data they have found on the web. This script should include text and character normalization and should standardize whitespace and capitalization rules. Another recommendation would be to restrict the length of text to be identical across all models and domains. A good starting point for such a script would be the punctuation normalizer from the Moses toolkit\footnote{\texttt{https://pypi.org/project/mosestokenizer/}} as this is what was used for the MAGE dataset \cite{li-etal-2024-mage}.

\paragraph{Include more variations across prompts and paraphrasers.} Prompts have been shown to wildly alter the stylistic components of generative model outputs even when only given task-oriented constraints---fooling detectors in the process \cite{koike-etal-2024-prompt}. In particular, experiments that test robustness to many different prompting strategies, paraphrase models, and synonym replacement methods are likely to give a strong sense of how well detectors will hold up in real-world settings. Strategies such as prefix-based prompting, length-conditioned generation, explicitly adversarial prompting, and others are all valid strategies to incorporate into future work.

\paragraph{Include more human-written text.} The imbalanced nature of the RAID dataset required participants to use data augmentation or up-sampling techniques that served only to degrade the quality of the data. Future work should seek to provide participants with a much larger corpus of human written texts from diverse domains. 

\end{document}